\documentclass[11pt]{article}

\usepackage[utf8]{inputenc}

\usepackage[top=1in, bottom=1in, left=1in, right=1in]{geometry}

\usepackage{epsfig, pdfpages}         
\usepackage{amsmath, amsfonts, amssymb}
\usepackage[shortlabels]{enumitem}
\usepackage{listings}
\usepackage{float}
\usepackage{array,multirow,graphicx}
\usepackage{subcaption}
\usepackage{amsmath,amsfonts,amssymb}
\usepackage{titling}
\usepackage{media9} 
\usepackage{hyperref}  
\usepackage{mathtools}
\usepackage{amssymb}
\usepackage{subcaption}
\usepackage{algorithm}
\usepackage{mathrsfs}
\usepackage[noend]{algpseudocode}

\providecommand{\keywords}[1]
{
  \small	
  \textbf{\textit{Keywords---}} #1
}

\title{Robust Trimmed $k$-means}

\begin{document}
\author{
Olga Dorabiala\footnote{email: olgad400@uw.edu},  J. Nathan Kutz, Aleksandr Aravkin \\[.1in]
{{\small Department of Applied Mathematics, University of Washington, Seattle, WA} }}

\date{\today}
\maketitle

\begin{abstract}
Clustering is a fundamental tool in unsupervised learning,  used to group objects by distinguishing between similar and dissimilar features of a given data set. One of the most common clustering algorithms is $k$-means. Unfortunately, when dealing with real-world data many traditional clustering algorithms are compromised by lack of clear separation between groups, noisy observations, and/or outlying data points.  Thus, robust statistical algorithms are required for successful data analytics.  Current methods that robustify $k$-means clustering are specialized for either single or multi-membership data, but do not perform competitively in both cases.  We propose an extension of the $k$-means algorithm, which we call {\em Robust Trimmed $k$-means} (RTKM) that simultaneously identifies outliers and clusters points and can be applied to either single- or multi-membership data.  We test RTKM on various real-world datasets and show that RTKM performs competitively with other methods on single membership data with outliers and multi-membership data without outliers.  We also show that RTKM leverages its relative advantages to outperform other methods on multi-membership data containing outliers. 
\end{abstract}




\section{Introduction}

Data science and machine learning have revolutionized the way that we do science today.  Intelligent systems are used in the engineering, physical, social, and biological sciences to take in data and output, among other critical information, actionable decision making capabilities or data analyses that detail correlations between important features ~\cite{bishop2006pattern,james2013introduction,goodfellow2016deep}.  The three major paradigms of machine learning are supervised, unsupervised, and reinforcement learning \cite{jordan2015machine}.  These three classes describe the kind of data used to structure learning tasks.  In {\em supervised learning}, the goal is to generate a learned mapping from inputs to outputs given labeled data.  The simplest example is linear regression, while a more complex example with high impact in recent years is deep neural networks \cite{schmidhuber2015deep,goodfellow2016deep}.  In contrast, {\em unsupervised learning} is used to discover the underlying patterns or structures of unlabeled data. This area includes methods for exploratory data analysis, such as dimensionality reduction and clustering.  Finally, {\em reinforcement learning} learns how to map situations to actions, so as to maximize a numerical (delayed) reward signal \cite{sutton2018reinforcement}. 
The relative successes of supervised and reinforcement learning are directly related to the availability of extensive labeled data.  In the absence of labels, these methods are known to perform poorly.  {\em Semi-supervised learning} attempts to address this issue by augmenting unlabeled data with smaller portions of labeled data \cite{goodfellow2016deep}.  However, it is often infeasible or expensive to manually label even a subset of a high-dimensional dataset.  In these cases, unsupervised learning techniques are the only available approach for extracting information.  Such methods are compromised by lack of clear separation between features, noisy observations, and/or outlying data points and require robustification, which is what we aim to improve in the context of the common $k$-means 
algorithm~\cite{bishop2006pattern,james2013introduction}.


Unsupervised learning techniques include dimensionality reduction, cluster analysis, and anomaly detection.  Although often treated as separate problems, these methods have significant overlap in practice.  Our specific algorithmic innovations pertain to the intersection between cluster analysis and anomaly detection.  Cluster analysis seeks to divide a set of objects so as to maximize both intra-cluster similarity and inter-cluster differences, while the aim of anomaly detection is to identify outliers in the dataset.      
Many diverse algorithms have been developed to solve these important problems~\cite{james2013introduction, bishop2006pattern}, including partitioning algorithms such as classic $k$-means and fuzzy $c$-means clustering ~\cite{askari2021fuzzy}, density based methods such as DBSCAN (density based spatial clustering of applications with noise)~\cite{ester1996density}, probabilistic methods such as mixture models~\cite{james2013introduction,bishop2006pattern}, and hiearchical clustering which produces dendrograms for data visualization~\cite{james2013introduction}.  In addition, spectral methods have been developed to extend the applicability of unsupervised learning to clusters that are not confined to spherical and/or elliptic distributions \cite{ng2002spectral}.






When data is well-separated and contains no outliers, $k$-means may be able to accurately assign labels to clusters \cite{bishop2006pattern}. 
Versions of the the $k$-means algorithm date back to the mid-1950s, with seminal contributions from Steinhaus~\cite{steinhaus1956division} and more modern versions developed by Lloyd~\cite{lloyd1982least} (published much later in 1982) and Forgy~\cite{forgy1965cluster} in the mid-1960s.  The $k$-means algorithm is simple, intuitive and can be directly applied without restrictions. Its simplicity and applicability have  contributed to its appeal and wide-spread usage; it was named one of the top-10 algorithms in data mining in 2008~\cite{wu2008top}.
Unfortunately, real-world data may be compromised by outliers and/or complicated by simultaneous membership to multiple clusters. Under these conditions, $k$-means is known to perform poorly.



The poor performance of machine learning algorithms on data with corruption and noise has long been acknowledged.  In the 1960s, John Tukey was the first to recognize the need for {\em robust} methods, coining the term {\em robust statistics}~\cite{huber2002john,donoho201750}.  Tukey was agnostic to any particular procedure, but simply insisted that working with real data required robustification in order to stabilize the performance and predictive power of machine learning and statistical methods.  Since that time, scientists have proposed numerous robustification techniques, including for clustering in unsupervised learning.  These methods include data trimming~\cite{rousseeuw2005robust, aravkin2020trimmed}, measures of outlierness~\cite{jiang2008clustering, he2003discovering, zhang2009new}, and staging methods for outlier identification~\cite{hautamaki2005improving}.
%
%
We propose a novel extension of the $k$-means algorithm, which we call {\em Robust Trimmed $k$-means} (RTKM), that allows us to (i) capture more information than previous methods, (ii) can be used to classify both single or multi membership data, and (iii) simultaneously clusters points and identifies outliers.  Experimental results show that RTKM, unlike other methods, performs competitively across all realms: on single-membership data containing outliers, multi-membership data without outliers, and multi-membership data containing outliers.

\section{Related Work}

In this section, we review existing approaches to robustify $k$-means clustering.  We focus on methods that build upon the basic $k$-means algorithm, due to $k$-means' simplicity, speed, and scalability \cite{bishop2006pattern}.  These developments also apply to any algorithm that depends on $k$-means, including spectral clustering, which can be applied to a variety of data distributions that $k$-means alone may not perform well on.  

We first review a fundamental connection between 
$k$-means clustering and optimization. 
The $k$-means algorithm can be viewed as an alternating minimization approach to solving the challenging optimization problem
\begin{equation}
    \min_{{\bf c},{\bf W}} \sum_{j=1}^k \sum_{i =1}^N w_{ji} ||{\bf x}_i - {\bf c}_j||^2 \text{  where  } \sum_{j=1}^k w_{j,i} = 1 \text{  for  } i = 1:N
\label{eqn:KMeansOpt}
\end{equation}
where ${\bf X} = [{\bf x}_1, \cdots {\bf x}_N] \in \mathbb{R}^{m \times N}$ are the data points and ${\bf C} = [{\bf c}_1, \cdots, {\bf c}_k] \in \mathbb{R}^{m \times k}$ are the cluster centers~\cite{huang2005automated}.  The matrix ${\bf W} \in \mathbb{R}^{k \times N}$ contains auxiliary weights $w_{j,i}$ that map the point-to-cluster relationship.  Each column $i$ of ${\bf W}$ assigns point ${\bf x}_i$ to a cluster whose center is ${\bf c}_j$.  If the weights are constrained to belonging to the discrete set  $w_{j,i} \in \{0,1\}$, problem~\eqref{eqn:KMeansOpt} is a mixed integer problem equivalent to classic clustering; it is nonsmooth and nonconvex.

\begin{figure}[t]
    \centering
    \includegraphics[width=.6\textwidth]{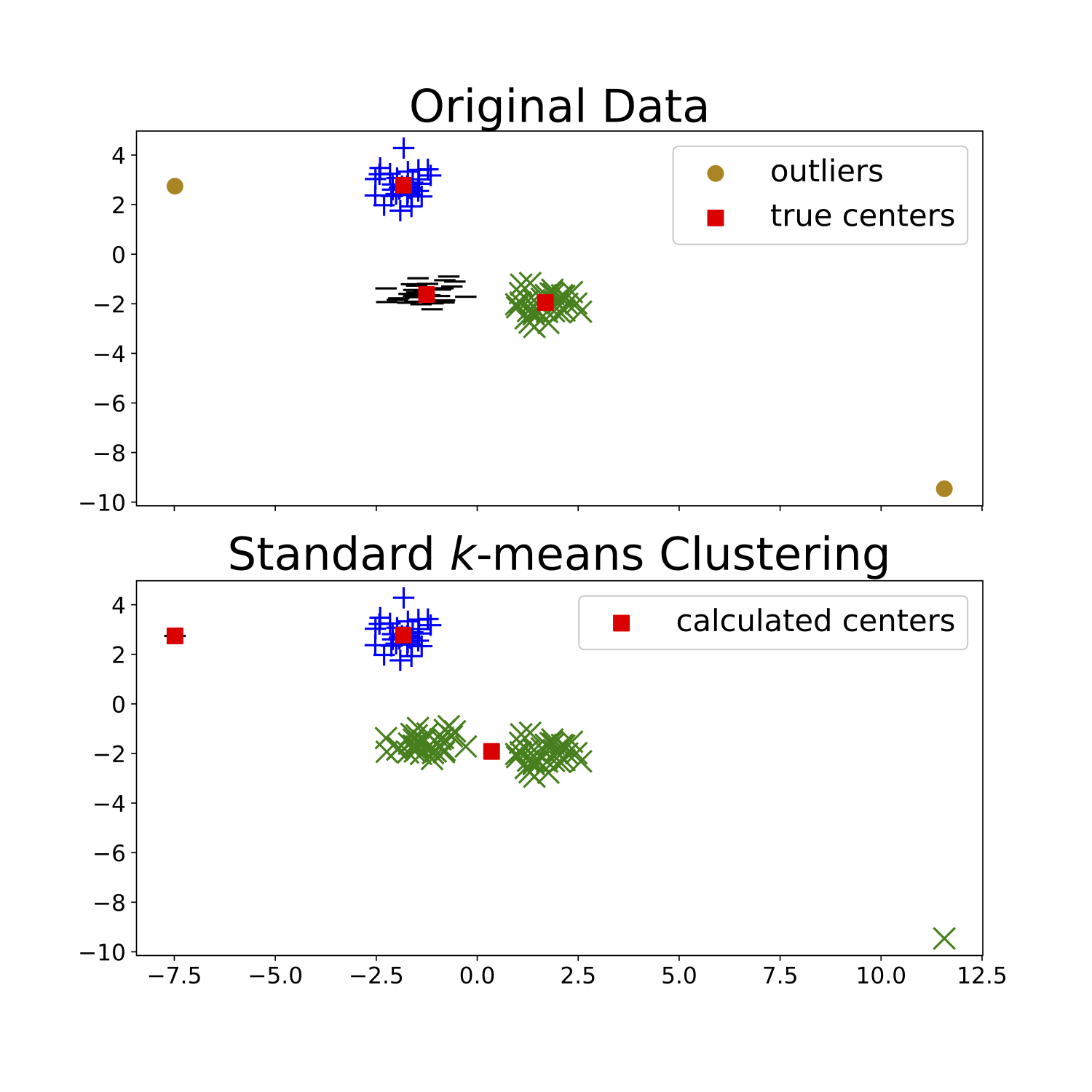}
    \caption{(top) Original data consisting of three clusters and two outliers. (bottom) Standard $k$-means incorrectly assigns a cluster center to the outlier on the right, causing two clusters to be misidentified as one.  The outlier on the left skews one cluster center assignment towards the bottom right.}
\label{fig:KMeans}
\end{figure}

The simplest approach to solving the  $k$-means problem is Lloyd's (1982) algorithm~\cite{lloyd1982least}.  Once cluster centers are (randomly) initialized, the algorithm works by alternatively assigning each point to its closest centroid and then updating cluster centers by taking the mean of all points in an assigned cluster.  This iterative process continues until convergence. This approach can be understood as a Gauss-Seidel type method for the nonsmooth, nonconvex problem~\eqref{eqn:KMeansOpt}. In each iteration, the variables are alternatively minimized as shown in \eqref{eqn:cupdate} and \eqref{eqn:wupdate} until convergence~\cite{huang2005automated}.  The
algorithmic updates in $k$-means are given below, with the pseudo-code shown in Algorithm \ref{alg:kmeans}
%
\begin{equation}
    {\bf c}_j^{k+1} = \frac{\sum\limits_{i=1}^N w_{j,i}^k {\bf x}_i}{\sum\limits_{i=1}^N w_{j,i}^k}
\label{eqn:cupdate}
\end{equation}
\begin{equation}
w_{ji}^{k+1} = \begin{cases}
1 & \text{if } ||{\bf x}_i - {\bf c}_j^{k+1}||^2 \leq ||{\bf x}_i - {\bf c}_t^{k+1}||^2 \text{ for } 1 \leq t \leq k \\
0 & \text{for } j\not=t
\end{cases}
\label{eqn:wupdate}
\end{equation}
While $k$-means works well in an ideal situation, one of its main drawbacks is sensitivity to outliers and noise.  Since points are classified by directly threshholding on the distance from cluster centers each iteration, outliers skew center assignment, and in turn point assignment, dramatically.  As shown in Figure \ref{fig:KMeans}, $k$-means is unable to properly classify points in the presence of outliers. 

\begin{algorithm}[t]
\caption{$k$-means}\label{alg:kmeans}

\begin{algorithmic}[1]
\Procedure{kmeans}{${\bf X},k$} \Comment{Input ${\bf X}= [{\bf x}_1, \cdots {\bf x}_n]$, $k$, and $d_k>1$}
\State Initialize ${\bf C} = [{\bf c}_1, \cdots {\bf c}_k]$
\While{not converged}
\State ${\bf c}_j^{k+1} = \frac{\sum\limits_i w_{ji}^k {\bf x}_i}{\sum\limits_i w_{ji}^k}$
\State $w_{ji}^{k+1} = \begin{cases}
1 & \text{if } ||{\bf x}_i - {\bf c}_j^{k+1}||^2 \leq ||{\bf x}_i - {\bf c}_t^{k+1}||^2 \text{ for } 1 \leq t \leq k \\
0 & \text{for } j\not=t
\end{cases}$

\EndWhile
\State ${\bf w}[:,i] = \arg \max_j {\bf w}[j,i]$
\State \textbf{return} ${\bf C}, {\bf w}$
\EndProcedure
\end{algorithmic}
\end{algorithm}

To solve this problem, numerous versions of a robust $k$-means algorithm have been proposed.  The majority of these methods perform clustering in stages.  In the first stage, the dataset is divided into clusters, and in the second stage, a measure based on the clusters is applied to the data to identify outliers.  One of the earliest examples of a staged method is trimmed $k$-means, proposed in 1997~\cite{cuesta1997trimmed}. Trimmed $k$-means works by running the standard $k$-means algorithm, removing a given percentage $\alpha$ of points with the greatest distance to their cluster centers, updating cluster centers as the mean of the remaining points in respective clusters, and repeating these steps until convergence.  Unfortunately, this method is ultimately ineffective, because it is unable to improve an already poor result by the standard $k$-means algorithm.  Other staging methods that suffer from the same drawback are those that define their own measures of outlierness, such as Outlier Removal Clustering (ORC)~\cite{hautamaki2005improving}, the cluster-based local outlier factor (CBLOF)~\cite{he2003discovering}, the outlier factor of a cluster~\cite{jiang2008clustering}, and Local Distance-Based Outlier Factor (LDOF)~\cite{zhang2009new}.  In using any of these methods, it is likely that points that should be classified as outliers are masked by the standard $k$-means clustering.  Therefore, a method that directly tackles outliers during clustering is preferred. 

Far fewer methods have been developed to simultaneously cluster and identify outliers.  Among those that do are Outlier Detection and Clustering algorithm (ODC)~\cite{ahmed2013novel}, $k$-means $-\,-$~\cite{chawla2013k},  Non-exhaustive, Overlapping $k$-means (NEO $k$-means)~\cite{whang2015non}, and $k$-means clustering with outlier removal (KMOR)~\cite{gan2017k}.  On single membership data containing outliers, KMOR outperforms all of the aforementioned methods in terms of cluster accuracy and outlier detection~\cite{gan2017k}.  KMOR always produces a single-membership assignment.  It identifies at most a given number of $n_0$ outliers by assigning them to a $k+1^{th}$ cluster based on whether they are further away than $\gamma$ times the average distance between inliers and their centroids.  The parameter $\gamma$ is difficult to identify when the proportion of outliers is unknown, and can drastically impact the results of the algorithm.  

Of all available methods, only NEO $k$-means, which extends $k$-means to overlapping clusters with noise, allows points to belong to multiple clusters simultaneously.  On multi-membership data without outliers, NEO $k$-means outperforms similar methods~\cite{whang2015non}.  The algorithm uses a single binary weight matrix to classify points and identify outliers, where parameters $\sigma$ and $\beta$ give the user a way to specify the degree of overlap and proportion of outliers, respectively.  Note that $\sigma$ is denoted as $\alpha$ in the original paper.  If a point is not assigned to any cluster, it is designated as an outlier.  NEO-$k$-means exhaustively assigns some multiple $(1 + \sigma)$ of the total number points to clusters, where $\sigma \geq 0$.  Because of this formulation, NEO-$k$-means is unable to identify outliers on datasets containing only a single cluster and cannot be restricted to single-membership on datasets containing outliers. 

We propose the Robust Trimmed $k$-means (RTKM) algorithm, which can be used to classify either single or multi-membership data in the presence of outliers and noise. In our numerical results, we therefore focus on the natural comparison between RTKM, KMOR, and NEO $k$-means.  We show that RTKM performs competitively with KMOR on single-membership data with outliers and with NEO $k$-means on multi-membership data without outliers. Moreover, RTKM leverages its advantages in these domains to
achieve superior performance on multi-membership data containing outliers.  In this way, RTKM can be effectively and competitively applied in multiple contexts.        
    
\section{Robust Trimmed k-means}

\subsection{Relaxation of k-means }
\label{sect:relax}
We propose an extension of the $k$-means objective in \eqref{eqn:KMeansOpt2} that allows points to belong to multiple clusters and sets up a foundation for a robust extension. Multi-cluster membership is appealing, because it allows one to identify the extent of a point's membership to every cluster~\cite{bezdek1980convergence}.  
Instead of restricting the auxiliary weight matrix ${\bf W}$ to the discrete set $\{0,1\}$, we allow ${\bf w}_{:,i} \in \Delta_s$ so that $\sum_{j=1}^k w_{j,i} =s$ for all $i$ and each $w_{j,i}$ is allowed to vary over the closed interval $[0,1]$.  The variable $s$ denotes the minimum number of clusters a point can belong to.  When $s=1$, the objective in \eqref{eqn:KMeansOpt2} is a classic relaxation of \eqref{eqn:KMeansOpt} where the weights $w_{j,i}$ can be interpreted as the probability that point $i$ belongs to cluster $j$. 
The relaxed objective is given by 
\begin{equation}
    \min_{{\bf c},{\bf W}} \sum_{j=1}^k \sum_{i =1}^N w_{ji} ||{\bf x}_i - {\bf c}_j||^2 \text{  where  } {\bf w}_{:,i} \in \Delta_s \text{  for  } i = 1:N, 
\label{eqn:KMeansOpt2}
\end{equation}
which is a continuous optimization problem, still nonsmooth and nonconvex. 
Problem \eqref{eqn:KMeansOpt2} can also be solved using alternating minimization.  The centers are updated as before in Algorithm \ref{alg:kmeans} using a Gauss-Seidel step. To better control how quickly the weights are updated, we use the Proximal Alternating Minimization (PAM) approach, which can be thought of as a proximal regularization of the Guass-Seidel scheme~\cite{attouch2010proximal}. 
The update for ${\bf W}$ is given in \eqref{eq:wupdate2}. PAM is guaranteed to converge as long as the step size $d_k>1$, and gives the practitioner additional control in managing the weight update in a rigorous way. 

\begin{equation}
 {\bf w}_{[:,i]}^{k+1} = \text{proj}_{\Delta_s} \bigg({\bf w}_{[:,i]}^k - \frac{1}{d_k}||{\bf x}_i - {\bf C}^{k+1}||^2\bigg) 
\label{eq:wupdate2}
\end{equation}

The pseudo-code for solving the relaxed $k$-means objective is shown in Algorithm \ref{alg:KmeansOpt} and allows for greater modeling flexibility.  In Algorithm \ref{alg:kmeans}, the columns of the weight matrix are projected onto the vertices of the capped simplex in step 5 of each iteration.  In Algorithm \ref{alg:KmeansOpt}, the weights can take on continuum of values in the entire capped simplex and are projected onto the vertices during cluster assignment in step 6 after the algorithm has already converged.  The relaxation allows Algorithm \ref{alg:KmeansOpt} to capture more information.  

\begin{figure}[t]
    \centering
    \includegraphics[width=.85\textwidth]{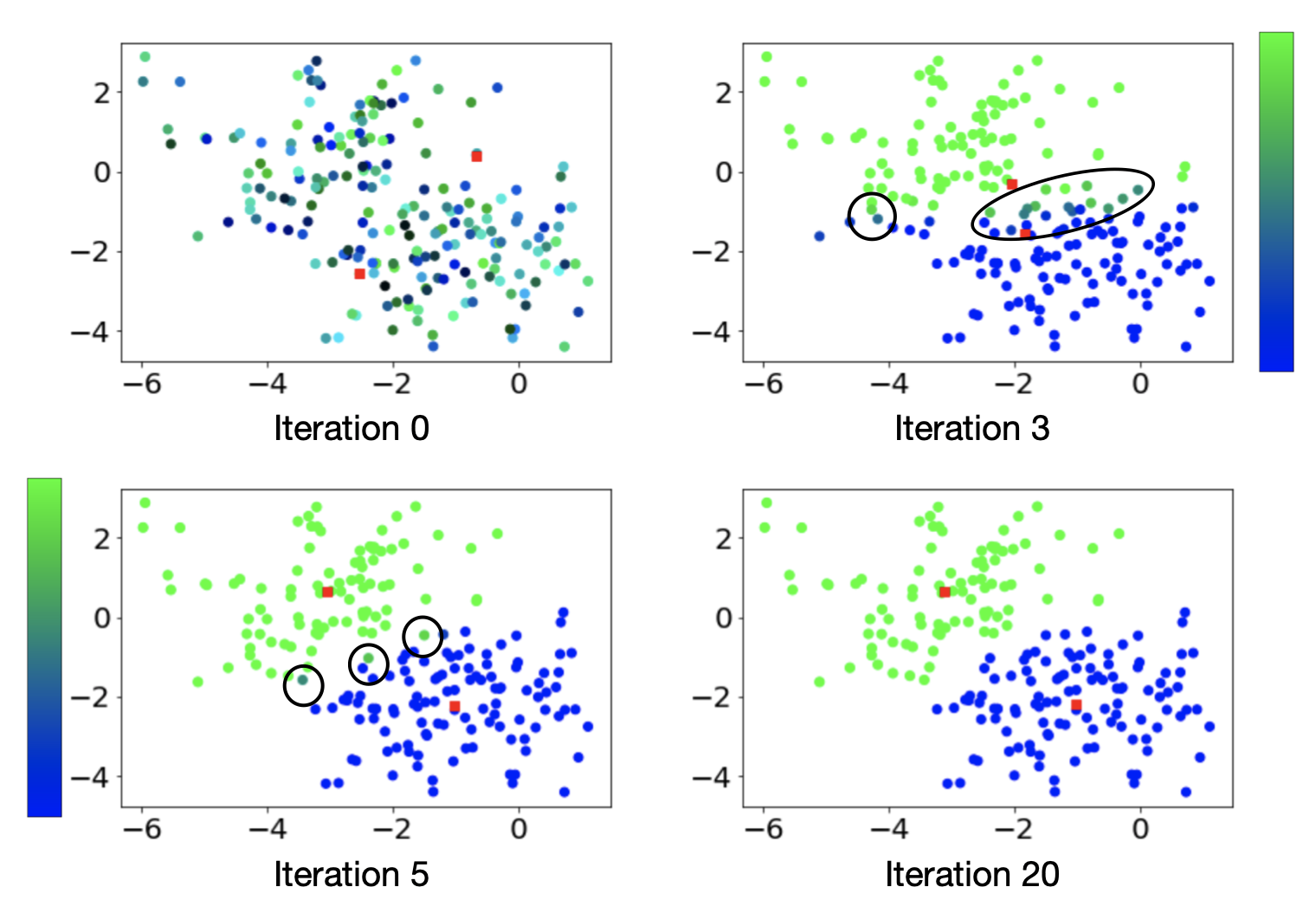}
    \caption{Clustering process of relaxed $k$-means.  The colors of the points represent the continuum of weights that assign points to clusters.  In iteration 0, all weights are assigned randomly and cluster centers are chosen randomly as well.  In iteration 3, two distinct clusters begin to form as the centers move.  In iteration 5, cluster centers begin to stabilize and points on the boundary retain partial membership to both clusters.  In iteration 20, relaxed $k$-means converges and two distinct clusters are formed.}
    \label{fig:clusteringProcess}
\end{figure}

Figure \ref{fig:clusteringProcess} gives a glimpse into how the clustering process works.  Point colors existing on the gradient between blue and green represent the continuous cluster weights.  In iteration $0$, the weights are randomly assigned and cluster centers are initialized at random.  By iteration 3, the centers begin to drift apart, and two clusters begin to form.  Points on the boundary remain on the gradient between blue and green.  In iteration 5, cluster centers begin to stabilize, and in iteration 20, relaxed $k$-means converges, and all the weights are on the vertices of the 1-capped simplex.  The relaxation of weighted $k$-means allows us to quantify the extent of membership of a point to each cluster at every iteration.

\begin{algorithm}[t]
\caption{relaxed $k$-means}\label{alg:KmeansOpt}

\begin{algorithmic}[1]
\Procedure{relaxedkmeans}{${\bf X},k, s$}
\Comment{Input ${\bf X=} [{\bf x}_1, \cdots {\bf x}_n]$, $k$, and $s$}
\State Initialize ${\bf C} = [{\bf c}_1, \cdots {\bf c}_k]$ and $d_k = 1.1$
\While{not converged}
    \State ${\bf c}_j^{k+1} = \frac{\sum\limits_i w_{ji}^k {\bf x}_i}{\sum\limits_i w_{ji}^k}$
    
    \State ${\bf w}_{:,i}^{k+1} = \text{proj}_{\Delta_1} \bigg({\bf w}_{:,i}^k - \frac{1}{d_k} ||{\bf x}_i - {\bf C}^{k+1}||^2\bigg)$
\EndWhile
\If{s = 1}
    \State ${\bf w}[:,i] = \arg \max_j {\bf w}[j,i]$
\Else 
    \State ${\bf w}[:,i] = \max({\bf w}[:,i], 0)$  
\EndIf
\State \textbf{return} ${\bf C}, {\bf w}$
\EndProcedure
\end{algorithmic}
\end{algorithm}

\subsection{Robust Trimmed k-means}

In order to create a robust $k$-means method, we model outlier detection using an analogous approach taken in Section \ref{sect:relax} for multi-cluster membership.  In this way, we avoid having to define our own measure of ``outlierness".  Our proposed method, which we call Robust Trimmed $k$-means (RTKM), simultaneously classifies points and identifies outliers by minimizing the objective function given in Equation \eqref{eq:MyMethodObj}.

\begin{equation}
\begin{split}
    \min_{{\bf c},{\bf v},{\bf W}} \sum_{i=1}^N v_i \sum_{j=1}^k w_{j,i} ||{\bf x}_i - {\bf c}_j||^2 \\
    {\bf w}[:,i] \in \Delta_s, \quad 
    {\bf v} \in \Delta_{N - [\alpha N]}
    \end{split}
\label{eq:MyMethodObj}
\end{equation}

\noindent As before, ${\bf X} = [{\bf x}_1, \cdots, {\bf x}_N] \in \mathbb{R}^{m\times N}$ are the data points, ${\bf C} = [{\bf c}_1, \cdots, {\bf c}_k] \in \mathbb{R}^{m\times k}$ are the cluster centers, and ${\bf W} \in \mathbb{R}^{k \times N}$ is a matrix of weights with each column in the s-capped simplex.  The new variable ${\bf v} \in \mathbb{R}^N$ is a vector identifying outliers that belongs to the $N - [\alpha N]$-capped simplex, where $\alpha$ is a given proportion of expected outliers.  Here $[\cdot]$ denotes the nearest integer, where we round up for half integer values.  The constraint on ${\bf  v}$ ensures that $N - [\alpha N]$ points are designated as inliers, since $\sum_{i=1}^N v_i = N - [\alpha N]$.  In the same way the constraint on the columns of ${\bf W}$ ensures that every point is assigned to at least $s$ clusters, since $\sum_{j=1}^k w_{j,i} = s$.  As discussed in Section \ref{sect:relax}, this constraint allows for each point ${\bf x}_i$ to belong to more than one cluster.  If we want to enforce single cluster membership, we do so by setting $s=1$ and make final assignments by calculating the $\arg \max$ over each column of ${\bf W}$ once the algorithm converges.    

The objective in Equation \eqref{eq:MyMethodObj} is approximately solved using Algorithm \ref{alg:MyAlg}.  We alternately minimize the objective with respect to all three variables: ${\bf W}, {\bf c}, \text{and } {\bf v}$.  We use a Gauss-Seidel update for the centers and Proximal Alternating Minimization (PAM)~\cite{attouch2010proximal} updates for both the columns of ${\bf W}$ and for ${\bf v}$, similar to the process described in Section \ref{sect:relax}.  Algorithm \ref{alg:MyAlg} is guaranteed to converge as long as $e_k >1$ and $d_k >1$.  In practice, we set $e_k= d_k = 1.1$.  

The algorithm takes as input the data matrix ${\bf X}$ and the number of clusters $k$.  Centroids can be initialized using any scheme.  The parameters $\alpha$ and $s$ control the number of outliers and number of members in each cluster, respectively.  If we know the percentage of outliers in a dataset, we can set $\alpha$ equal to that value.  Likewise, if we know the cardinality of our dataset, we can set $s = \left \lfloor{\text{cardinality}}\right \rfloor$.  Currently, there is no principled approach to estimating these paramenters.  However, we demonstrate that RTKM performs competitively with other methods on data sets containing outliers using a range of $\alpha$ values.  

\begin{algorithm}[t]
\caption{Robust Trimmed $k$-means (RTKM)}\label{alg:MyAlg}

\begin{algorithmic}[1]
\Procedure{RTKM}{${\bf X},k, \alpha, s$} \Comment{Input ${\bf X}= [{\bf x}_1, \cdots {\bf x}_n]$, $k$, $\alpha$, and $s$}
\State Initialize ${\bf C} = [{\bf c}_1, \cdots, {\bf c}_k]$, $e_k = 1.1$, and $d_k = 1.1$
\While{not converged}
\State ${\bf c}^{k+1}[:,j] = \frac{\sum\limits_{i=1}^N v_i^k w_{ji}^k {\bf x}_i}{\sum\limits_{i=1}^N v_i^k w_{ji}^k}$ 

\State ${\bf w}_{[:,i]}^{k+1} = \text{proj}_{\Delta_s}\bigg({\bf w}_{[:,i]}^k - \frac{1}{d_k}v_i^k||{\bf x}_i - {\bf C}^{k+1}||^2\bigg)$

\State $v^{k+1} = \text{proj}_{\Delta_{[\alpha N]}} \bigg( v^k - \frac{1}{e_k}\sum_{j=1}^k {\bf w}_{[j,:]^{k=1}} ||{\bf X} - {\bf c}_j^{k+1}||^2 \bigg)$

\EndWhile

\If{s = 1}
    \State ${\bf w}[:,i] = \arg \max_j {\bf w}[j,i]$
\Else 
    \State ${\bf w}[:,i] = \max({\bf w}[:,i],0)$
\EndIf
\State \textbf{return} ${\bf C}, {\bf w}, {\bf v}$
\EndProcedure
\end{algorithmic}
\end{algorithm}

\section{Experiments}

We compare the performance of RTKM against other methods that simultaneously perform clustering and outlier detection, specifically KMOR \cite{gan2017k} and NEO-$k$-means \cite{whang2015non}.  We focus our comparison against KMOR and NEO-$k$-means, because these methods were shown to outperform others on various datasets.  In their paper, Gan et.al \cite{gan2017k} showed that KMOR outperformed ODC \cite{ahmed2013novel}, $k$-means$--$ \cite{chawla2013k} and NEO-$k$-means \cite{whang2015non} on single-membership data containing outliers.  Similarly, NEO-$k$-means \cite{whang2015non} was shown to outperform MOC~\cite{banerjee2005model}, fuzzy $k$-means~\cite{bezdek1980convergence}, explicit sparsity constrained clustering (esp)~\cite{lu2012overlapping}, implicit sparsity constrained clustering (isp)~\cite{lu2012overlapping}, OKM~\cite{cleuziou2008extended}, and restricted OKM (rokm)~\cite{ben2013identification} on multi-membership data without outliers.  We evaluate the performance of RTKM on three types of datasets: single-membership datasets containing outliers, multi-membership datasets without outliers, and mutli-membership datasets with outliers.

The quality of the cluster assignments is measured using the metric in \cite{whang2015non}, average $F_1$ score, which quantifies how well each algorithm finds the ground truth clusters.  The $F_1$ score is defined as, 

\begin{equation}
    F_1 = \frac{TP}{TP + \frac{1}{2} (FP + FN)}
\end{equation}

\noindent where $TP$ denotes true positives, $FP$ denotes false positives, and $FN$ denotes false negatives. This metric ranges from $0$ to $1$, with values closer to $1$ implying better classification.  To calculate the average $F_1$ score, predicted clusters are matched to ground-truth clusters so that the average $F_1$ score among all clusters is maximized.  On datasets containing outliers, outliers are considered their own cluster for the purpose of calculating the average $F_1$ score. 

The ability to correctly identify outliers is measured using the distance of a classifer on the Receiver Operating Curve graph from the perfect outlier classifier ($M_e$) as in \cite{gan2017k}.  The measure $M_e$ is defined as,

\begin{equation}
    M_e = \sqrt{(FP_{\text{rate}})^2 + (1 - TP_{\text{rate}})^2}
\end{equation}

\begin{equation}
    TP_{\text{rate}} = \frac{TP}{TP + FN} 
\end{equation}

\begin{equation}
FP_{\text{rate}} = \frac{FP}{FP + TN}
\end{equation}

\noindent where $TP$, $FP$, and $FN$ are defined as before and $TN$ denotes true negatives.  The $M_e$ score depends only on the true and identified outliers in the dataset.  The value of $M_e$ ranges from $0$ to $\sqrt{2}$, with better outlier classifiers having values closer to $0$.  

\subsection{Single-membership Data with Outliers}\label{subsect:Outliers}

We begin by evaluating RTKM against KMOR and NEO-$k$-means on two single-membership datasets containing outliers.  The first is the Breast Cancer Wisconsin (WBC) dataset \cite{mangasarian1990cancer} from UCI Machine Learning Repository \cite{Dua:2019}.  The WBC dataset contains $699$ instances of tumors with $9$ numerical attributes each.  All of the instances are classified as either benign or malignant, of which we treat the latter as outliers.  
 
\begin{figure}[t]
    \centering
    \includegraphics[width=.50\textwidth]{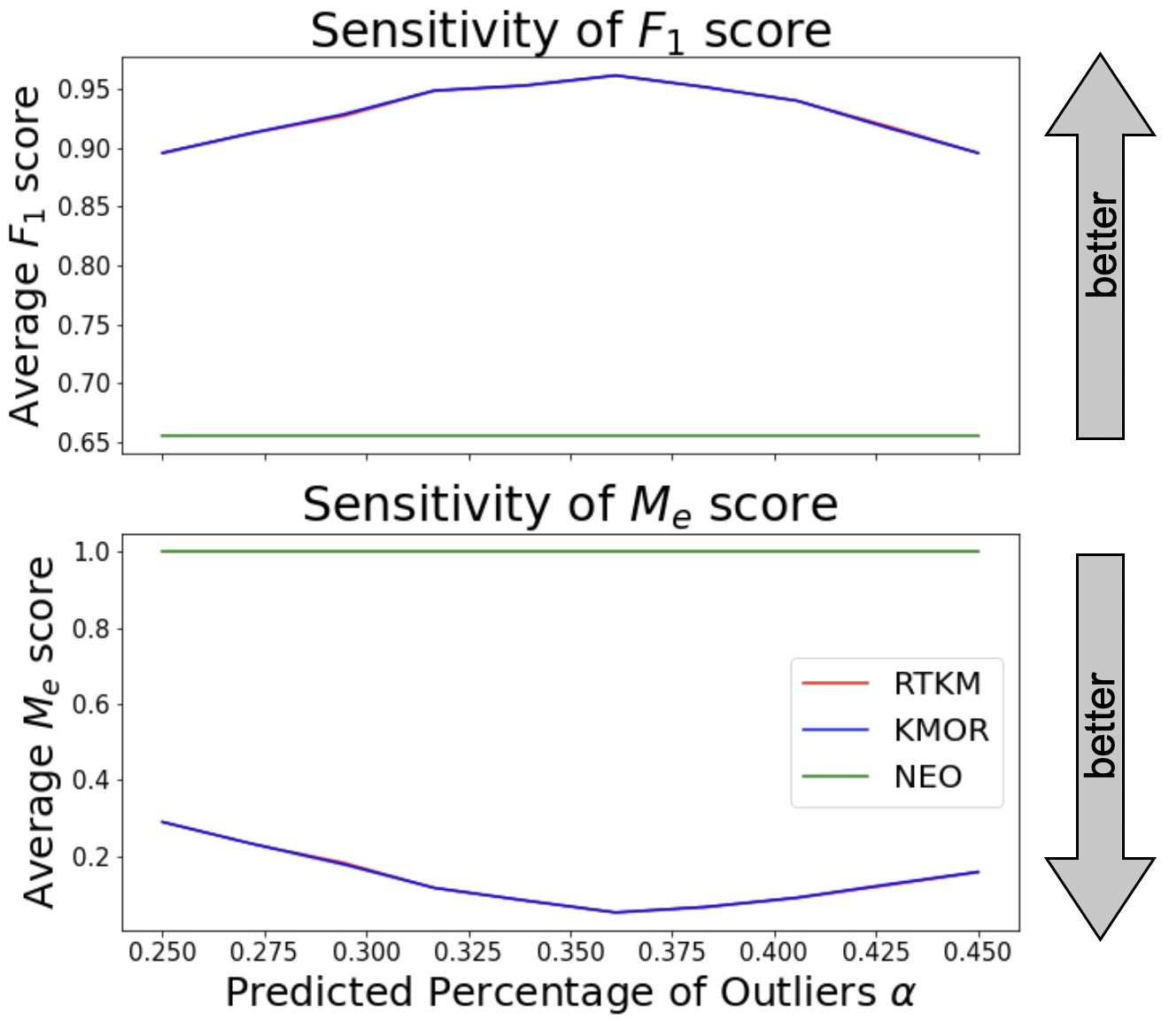}
    
    \caption{Sensitivity of RTKM, KMOR, and NEO-$k$-means to the choice of expected percentage of outliers on the WBC dataset.  RTKM and KMOR exhibit almost identical performance as measured by $M_e$ and $F_1$ scores. NEO-$k$-means is unable to identify any outliers on the WBC dataset.}
\label{fig:sensitivityWBC}
\end{figure}

 All three methods are set to search for $k=1$ clusters, with one member belonging to each cluster.  Various values are used for the expected percentage of outliers $\alpha$ to see how sensitive the results are to parameter choice.  For every value of $\alpha$, we complete 10 runs of each algorithm with different cluster center initialization each time.  Performance metrics are reported as the minimum, maxmimum, and average $M_e$ and $F_1$ scores over the 10 runs.  Figure \ref{fig:sensitivityWBC} shows these performance metrics for the three algorithms over a range of $\alpha$ values on the WBC dataset.  RTKM and KMOR exhibit almost identical performance when given the same parameters, while NEO-$k$-means is unable to identify any outliers, regardless of parameter choice.  This is because, by design, when $k=1$, NEO-$k$-means cannot identify outliers. 
 
Next, we test the three algorithms on the shuttle training dataset, also from the UCI Machine Learning Repository \cite{Dua:2019}.  This dataset contains 43,500 records and 7 classes, described by 9 numerical features.  The three largest classes contain $99.57\%$ of the data.  Therefore, we consider the three largest classes as inliers, and the remaining four classes to be outliers.  

\begin{figure}[t]
    \centering

\includegraphics[width=.5\textwidth]{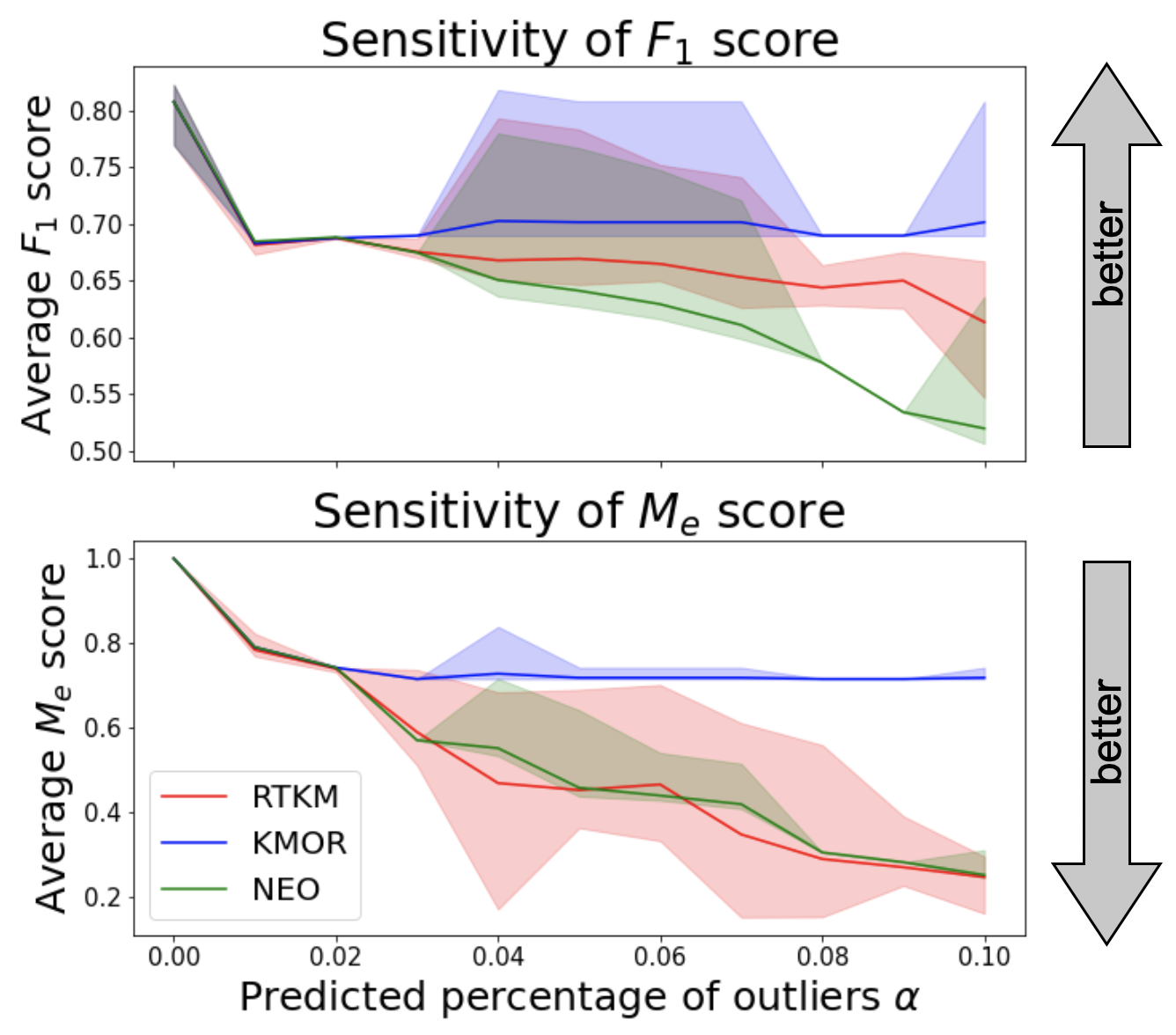}

    \caption{Sensitivity of RTKM, KMOR, and NEO-$k$-means to choice of expected number of outliers on the Shuttle dataset.  All three methods perform similarly until the predicted percentage of outliers surpasses $0.02$. Then, KMOR and NEO-$k$-means identify significantly more outliers correctly, but the corresponding increase in false positives leads to lower $F_1$ scores. }
\label{fig:sensitivityshuttle}
\end{figure}

We set $k=3$ as the number of clusters and allow each point to belong to at most one cluster.  As before, we test the sensitivity of the results for RTKM, KMOR, and NEO-kmeans to the choice of expected percentage of outliers in the dataset.  Figure \ref{fig:sensitivityshuttle} shows minimum, maximum, and average $M_e$ and $F_1$ scores for 10 runs of each of the methods for various $\alpha$ values.  Up until the predicted percentage of outliers becomes $0.02$, all three methods perform similarly.  Beyond this point, increasing $\alpha$ leads KMOR and NEO-$k$-means to identify more outliers correctly and achieve much lower $M_e$ scores compared to KMOR.  However, the corresponding increase in false positive outliers leads to comparatively lower $F_1$ scores.  In applications where there is a high cost associated with false positive outliers, KMOR may be preferential.  Conversely, in applications were the goal is to identify as many outliers as possible, RTKM is superior.     

\subsection{Multi-membership Data without Outliers}\label{sect:clustoverlap}

Unlike KMOR, both RTKM and NEO-$k$-means have the ability to classify points as belonging to multiple clusters \cite{whang2015non}.  The parameters in NEO-$k$-means determine exactly how many point assignments are made and at most how many outliers can be identified.  In contrast, RTKM's parameters determine at least how many assignments are made and at least how many outliers are identified.  In their recent paper, Whang et.al \cite{whang2015non} showed that NEO-$k$-means outperformed six other methods that study multi-cluster membership on two real datasets.  We evaluate RTKM and KMOR on the same two datasets and add their performance, as measured by average $F_1$ score to the evaluation from \cite{whang2015non}.  

We use the ``yeast" and ``scene" datasets from \cite{mulan}.  The ``yeast" dataset contains $2417$ instances with $103$ numerical attributes.  There are $14$ classes in total, and the cardinality of the dataset is $4.237$.  The ``scene" dataset contains $2407$ instances with $294$ numeric attributes.  There are $6$ classes in total, and the cardinality of the dataset is $1.074$.  Both datasets are also used for testing MOC~\cite{banerjee2005model}, fuzzy $k$-means~\cite{bezdek1980convergence}, explicit sparsity constrained clustering (esp)~\cite{lu2012overlapping}, implicit sparsity constrained clustering (isp)~\cite{lu2012overlapping}, OKM~\cite{cleuziou2008extended}, and restricted OKM (rokm)~\cite{ben2013identification}.  We borrow performance reports for all algorithms except RTKM and KMOR from \cite{whang2015non}.

For the ``yeast" dataset, we use $k = 14$, $\alpha = 0$, and $s= 4$ as our inputs for RTKM and $k = 14$, $n_0 = 0$, $\gamma = 9$ as our inputs for KMOR.  For the ``scene" dataset, we use $k = 6$, $\alpha = 0$, and $s = 1$ as our inputs for RTKM and $k = 6$, $n_0 = 0$, $\gamma = 9$ as our inputs for KMOR.  Each algorithm is run five times using the same cluster center initialization, and the result that leads to the best objective function value for each method is chosen, as in~\cite{whang2015non}.  Average $F_1$ scores for each algorithm on both datasets are shown in Table \ref{tab:overlap}.  NEO-$k$-means achieves the highest $F_1$ score on both datasets.  However, in both cases the $F_1$ score of RTKM is only slightly lower.  In fact, it is the second highest of all the scores reported.  On the ``scene" dataset, KMOR performs just as well as RTKM.  This result is due to the cardinality of the ``scene" dataset being so close to one.  Contrastingly, on the ``yeast" dataset, KMOR is unable to produce a competitive result due to a higher cardinality. 

Recall that the optimal choice for the parameter in RTKM controlling the number of clusters each point is assigned to is $s = \left \lfloor{\text{cardinality}}\right \rfloor$.  This means that if the cardinality of a dataset is only slightly larger than one, RTKM will perform better doing single-, rather than multi-, cluster membership.  RTKM's success on multi-membership data is most pronounced on datasets where $s\geq 2$.

\begin{table}[t]
    \centering
    \begin{tabular}{|c|c|c|c|c|c|c|c|c|c|}
    \hline
        & RTKM & KMOR & NEO-$k$-means & moc & fuzzy & esp & isp & okm & rokm \\
        \hline
         ``yeast" & 0.317  & 0.161 & 0.366 & - & 0.308 & 0.289 & 0.203 & 0.311 & 0.203 \\
        \hline
         ``scene" & 0.597 & 0.597 & 0.626& 0.467 & 0.431 & 0.572 & 0.586 & 0.571 & 0.593\\
         \hline
    \end{tabular}
    \caption{Average $F_1$ scores of various multi-membership clustering methods on the ``yeast" and ``scene" datasets.  NEO-k-means achieves the best $F_1$ score on both datasets.  However, RTKM achieves the second highest $F_1$ score on both datasets, demonstrating its competitiveness.}
    \label{tab:overlap}
\end{table}

\subsection{Multi-membership Data with Outliers}

In order to compare RTKM against KMOR and NEO-$k$-means on multi-membership data containing outliers, we add noise to the ``yeast" dataset from Section \ref{sect:clustoverlap}.  We do so by adding $150$ noise points, so that the data contains $\sim 10\%$ outliers.

\begin{figure}[t]
    \centering
    \includegraphics[width=.50\textwidth]{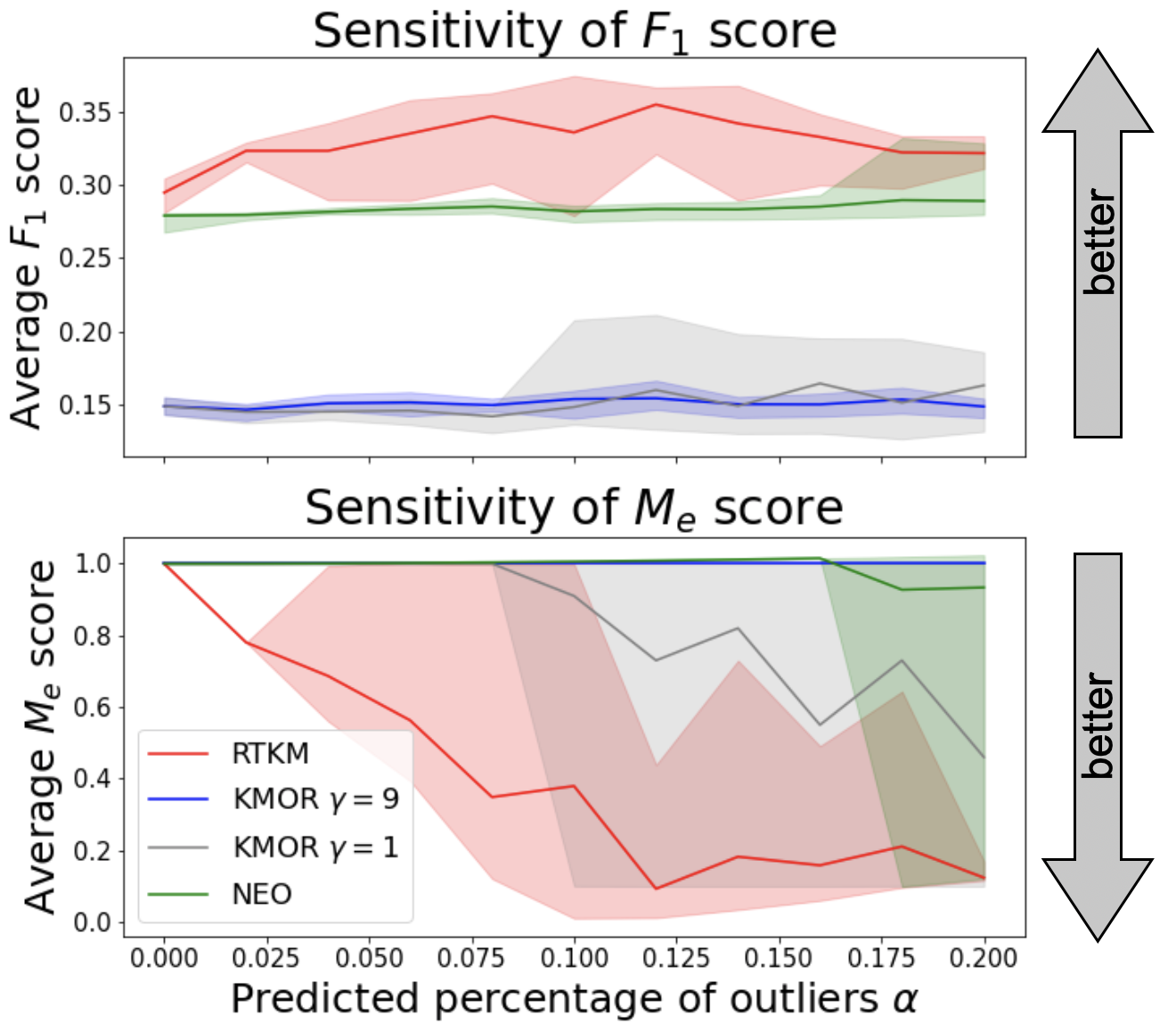}
    \caption{Sensitivity of RTKM, KMOR, and NEO-$k$-means to the choice of expected percentage of outliers on the Yeast plus noise dataset. Over all parameter choices, RTKM achieves the highest average $F_1$ score and the lowest average $M_e$ score.}
    \label{fig:yeastsummary}
\end{figure}

We set $k=14$ as the number of clusters.  Additionally, RTKM is given $s=4$ so that each point is assigned to at least $4$ clusters, and NEO-$k$-means is given $\sigma = 3$, so that each point is assigned to at most $4$ clusters.  KMOR exhibits different performance when using $\gamma =1$ versus $\gamma = 9$, so we include results for both initializations.  We again test the sensitivity of the results of all three methods to the expected percentage of outliers $\alpha$.  Figure \ref{fig:yeastsummary} shows the minimum, maximum, and average $M_e$ and $F_1$ scores for 10 runs of each of the methods using various parameter values.   

\begin{figure}[t]
    \centering
    \includegraphics[width=\textwidth]{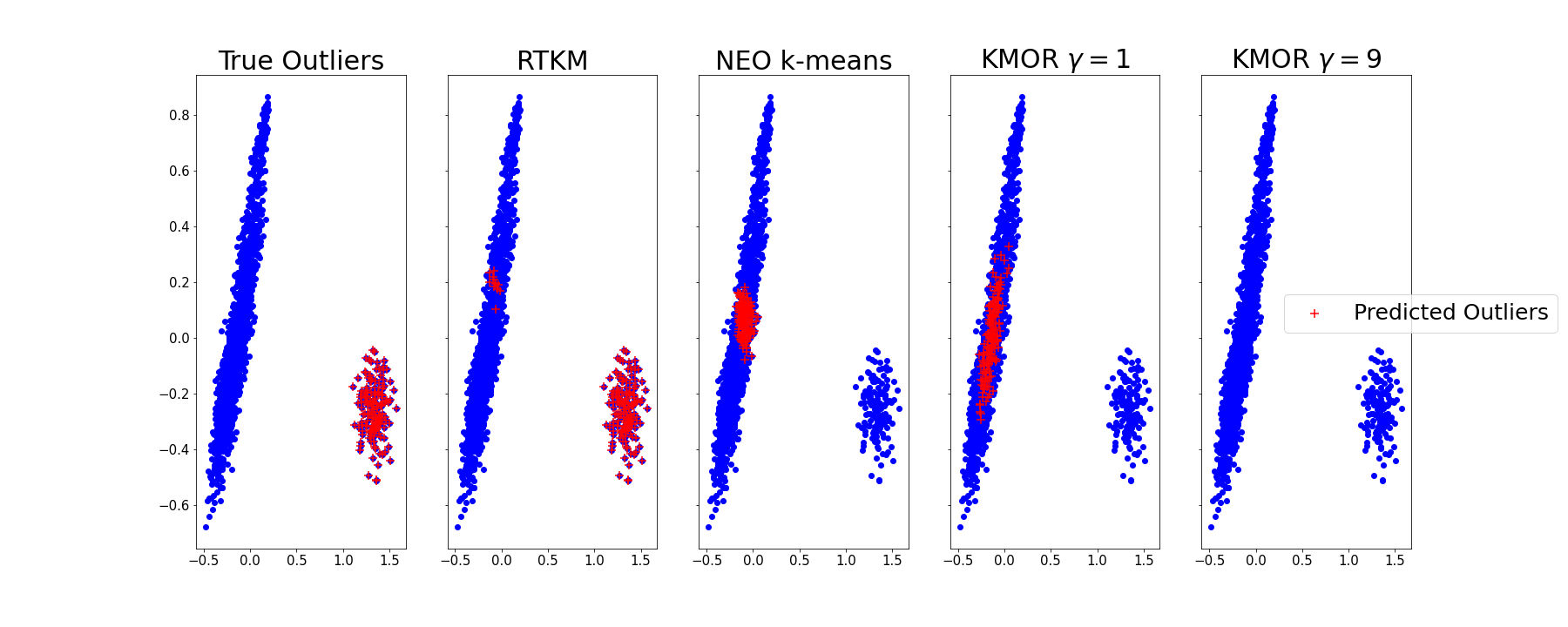}
    \caption{Outliers as identified by RTKM and NEO-$k$-means on the ``yeast" dataset with noise.  The leftmost figure shows the true location of outliers.  The remaining figures show the location of outliers identified by RTKM, NEO-$k$-means, KMOR with $\gamma = 1$, and KMOR with $\gamma =9$ for the same cluster center initialization.}
    \label{fig:yeastoutlier}
\end{figure}

As seen in Figure \ref{fig:yeastsummary}, of the three methods tested, RTKM achieves the highest average $F_1$ and the lowest average $M_e$ scores.  When $\gamma = 9$, KMOR does not identify any outliers, regardless of parameter choice.  Given a value of $\alpha$ greater than $0.100$, KMOR with $\gamma = 1$ and NEO-$k$-means are sometimes able to achieve almost as low of an $M_e$ score as RTKM, but on average, given the same initialization, RTKM does a better job at identifying outliers.  Figure \ref{fig:yeastoutlier} showcases this behavior.  Given the same cluster center initialization, RTKM is able to correctly identify all of the outliers plus a few false positives, whereas NEO-$k$-means and KMOR do not identify any outliers correctly.  This demonstrates that on multi-membership data containing noise, RTKM is able to leverage its relative advantages over KMOR and NEO-$k$-means to outperform both methods. 

\section{Conclusions and Future Work}

We propose Robust Trimmed $k$-means (RTKM) as an algorithm for simultaneous point classification and outlier detection that can operate on both single- and multi-membership data.  In RTKM, the parameters $k$ and $\alpha$ control the number of clusters and expected percentage of outliers respectively.  The parameter $s$ controls at least how many clusters we want each point to belong to.  At the moment, there is no principled approach to estimating these parameters.  We leave the investigation of such an approach for future work, and demonstrate that for the time being, RTKM remains competitive with existing methods over a range of $\alpha$ values. 

The innovations presented rely on a robust relaxed formulation for the weighted $k$-means algorithm that allows the classification weight matrix to exist on a continuum of values $[0,1]$, rather than the binary set $\{0,1\}$.  This relaxation gives the user a way to track the extent of membership of a point to each cluster at every iteration and provides flexibility for multi-cluster membership.  We apply the same methodology for outlier identification, thereby avoiding explicitly defining a measure of ``outlierness", unlike many other methods. Relaxation-based formulations have proven to be effective in a number of recent applications~\cite{zheng2018unified,champion2020unified}.
In the context of the current application, relaxation to a continuum of values, coupled with the PAM algorithm, provides a way to search the model space more effectively in order to discover clusters and outliers. Using this method, we avoid rushing into a bad local minimum, as with prior alternating methods moving across extreme points of the simplex.

We test RTKM on three types of datasets: single-label data with outliers, multi-label data without outliers, and mutli-label data with outliers.  While KMOR and NEO-$k$-means set the benchmark for the first two types of datasets, respectively, they do not perform well on both.  RTKM remains competitive on both types of data, and outperforms both KMOR and NEO-$k$-means on multi-label data with outliers.  On single-label data, RTKM performs almost identically to KMOR in terms of $F_1$ and $M_e$ scores on a simple dataset containing one cluster with outliers.  NEO-$k$-means produces a poor result since by design, the method cannot identify outliers in a dataset containing only one cluster.  On a single-membership dataset containing multiple clusters with outliers, RTKM outperforms NEO-$k$-means, but achieves a lower $F_1$ score than KMOR.  Even so, both RTKM and NEO-$k$-means identify more outliers correctly than KMOR, as evidenced by lower $M_e$ scores.  We conclude that in applications where the cost of false positive outliers is high, KMOR may be preferred, but in applications where such a cost is low and it is important to identify as many outliers as possible, RTKM may be favored.  On multi-label data without outliers, RTKM remains competitive against NEO-$k$-means, achieving a higher $F_1$ score than every other method NEO-$k$-means is compared against in \cite{whang2015non}.  Furthermore, RTKM substantially outperforms KMOR when the cardinality of a dataset is greater than or equal to $2$, since KMOR does not have the functionality to make multi-membership assignments.  On multi-label data containing outliers, RTKM achieves the highest average $F_1$ score and the lowest average $M_e$ score across all tested values for predicted percentage of outliers.  On average, given the same cluster center initialization, only RTKM is able to correctly identify any outliers.  These experiments demonstrate that although RTKM has a niche where it performs best, it does so without sacrificing performance on other types of datasets. \\

\noindent
{\bf Code Availability}: \url{https://github.com/OlgaD400/Robust-Trimmed-K-Means} \\

\bibliographystyle{abbrv}
\bibliography{mybib}

\end{document}